# Compressive Representations of Weather Scenes for Strategic Air Traffic Flow Management

Sandip Roy
School of Electrical Engineering and Computer Science
Washington State University
Pullman, WA
sandip@wsu.edu

*Abstract*—Terse representation of high-dimensional weather scene data is explored, in support of strategic air traffic flow management objectives. Specifically, we consider whether aviation-relevant weather scenes are compressible, in the sense that each scene admits a possibly-different sparse representation in a basis of interest. Here, compression of weather scenes extracted from METAR data – including temperature, flight categories, and visibility profiles for the contiguous United States – is examined, for the graph-spectral basis. The scenes are found to be compressible, with 75-95% of the scene content captured using 0.5-4% of the basis vectors. Further, the dominant basis vectors for each scene are seen to identify time-varying spatial characteristics of the weather, and reconstruction from the compressed representation is demonstrated. Finally, potential uses of the compressive representations in strategic TFM design are briefly scoped.

## I. BACKGROUND AND INTRODUCTION

Air traffic flow management (TFM) aims to shape traffic demand to match capacities of airspace resources (e.g., airports, Sectors), without imposing extraneous delay and cost [1,2]. Because airspace resource capacities are principally dependent on weather [3,4], forecasting weather and mitigating its impacts on airspace system performance are foundational challenges in TFM. For strategic TFM, which is concerned with airspace-system-wide management at lookaheads of several hours to a day, challenges remain in weather forecasting and automated TFM design [2,5,6]. Indeed, strategic TFM is still primarily developed manually relying on operators' experience, rather than based on forecast- and data-based decision aids. While this manual paradigm has generally worked well, it is likely to become untenable as the airspace becomes increasingly heterogeneous and sophisticated, and traffic demands increase. Thus, there is considerable interest in developing decision aids and automation for strategic TFM.

In recent years, several approaches have been proposed for designing strategic TFM in the face of severe weather [2,5-10]. Most of these approaches can be roughly categorized as being either traffic-model-based [2,5-8], or historical-comparison-based [9,10]. Both types of approaches depend on forecasts of weather and its impact on airspace capacities, as well as uncertainty quantifications for these forecasts. With this motivation, techniques have been proposed to translate standard weather forecasts (e.g., ensemble forecasts) to aviation-relevant forecasts (e.g., of flight categories/rules or Sector capacities) [11-15]. Building on these forecasting tools, the model-based methods for TFM design use simulations of air traffic models under forecasted weather to predict system performance, and in turn tune traffic management initiatives (TMIs) to shape the performance. Meanwhile, the historical-comparison-based approaches compare weather futures with historical scenarios, and use these comparisons to suggest traffic management plans.

The historical-comparison-based approaches often require identification of key features or low-dimensional bases for weather, which can be compared against historical scenarios to select TMIs. Likewise, model-based design approaches require knowledge of key weather features which impact traffic bottlenecks, since focusing on key features can allow pruning of the design space and enable learning/optimization without overfitting. Thus, many of the proposed methods used for planning strategic TFM require terse representations of airspace-relevant weather scenes (profiles).

Efforts to represent weather data in a terse form have primarily focused on identifying a set of key components or bases for a historical data ensemble. For example, Mukherjee et al have used principal component analysis (PCA) and factor analysis to identify dominant components in en route impacts of severe weather [9,16,17]. In analogy, in comparing airport weather profiles with the historical archive, Buxi and Hansen have also proposed using PCA followed by K-means clustering to partition the archived data based on key features [10]. These various approaches, at their essence, presuppose that airspace-relevant weather profiles follow some typical patterns, which can be represented as a linear combination of a small set of basic vectors. However, weather and weather impact profiles in reality may not be representable using a small set of fixed basis vectors, as weather events exhibit huge variability in their spatial extent, pattern, severity, and evolution.

In this study, we explore an alternative, less-restrictive approach for terse representation of aviation-relevant weather profiles or scenes. Specifically, we study whether the scenes are *compressible*, in the sense that each scene in an ensemble or archive can be expressed as a possibly-different sparse combination of basis vectors in a specified full basis [18-20]. The concept underlying our approach is that, while weather profiles/scenes vary greatly over time, each profile also exhibits considerable structure such as strong spatial correlation, periodicities, typical intensity patterns, etc. Thus, in an appropriately selected full basis, it is reasonable that each ensemble member can be expressed in terms of a different small set of basis vectors. This compressibility concept for terse data representation is the brainchild of the statistical signal processing and particularly the compressive sensing community, where it is used for reconstruction of signals from sparsely-sampled data [18-20]. The compressibility concept has found application in many domains, ranging from seismology (where data is often compressible in the Fourier basis because measurements have content at only a few tones or frequencies) and image processing.

The main focus of this article is to explore whether continental-scale weather data that is relevant for strategic traffic flow management is compressible. Specifically, compressibility is studied for TFM-relevant weather scenes derived from METAR data from across the continental United States. Specifically, in this initial study, three quantities are considered: 1) temperature, 2) flight category/rule, and 3) visibility. Because these weather data are expected to exhibit spatial patterns, compression in a basis that reflects the spatial structure of the measurements is considered. Due to the irregular sampling provided by METARs, we consider a basis defined from a graph representing the METAR locations. In particular, compressibility in the graph spectral or Laplacian-eigenvector is considered. Our analysis shows that the scene data is compressible, with accurate representations of the ~2500 data points captured using only 10-100 basis vectors. After this main effort, we also biefly discuss ways in which compressibility can be brought to bear to support strategic traffic flow management.

The article is organized as follows. Compressibility notions are defined in Section II. Then, the method used for compressing airspace-relevant weather scenes is explained in Section III. Results are shown for several case study data sets in Section IV. Finally, implications of compressibility on strategic TFM are briefly discussed in Section V.

II. COMPRESSIBILITY: DEFINITIONS AND CONCEPTS

The notions of compression and compressibility considered in this work have to do with terse representation of a vector $x$ with $N$ entries, which is drawn stochastically from an ensemble $X$. In our setting, the vector $x$ represents an aviation-relevant weather scene, i.e. a snapshot of a weather quantity of interest sampled at locations across the airspace. We are interested in sparse representation of the vector $x$ in a particular $N$-dimensional orthogonal basis, which is specified by the columns of an $N \times N$ matrix $F$. The notions of compressibility, defined next, closely follow the literature on compressive sensing [18-22].

The vector $x$ is said to be *exactly K-compressible* in the basis $F$, if $x = Fs$ for some vector $s$ which has at most $K$ non-zero entries. The ensemble $X$ is said to be *exactly K-compressible,* if every vector in $X$ is $K$-compressible.

In practice, data is rarely exactly compressible in any basis, and approximate notions of compressibility are also needed. Conceptually, the vector can be viewed as approximately compressible, if a sufficient fraction of the energy in the vector is captured by a sparse approximation in the basis of interest. Specifically, we refer to the maximum fraction of the energy in the vector $x$ captured by a linear combination of $K$ basis vectors as the *K-compressibility level* of the vector. Mathematically, the $K$-compressibility level $L$ can be expressed as:

$$L = 1 - \frac{\min_{s \text{ s.t. } ||s||_0 = K} ||x - Fs||_2^2}{||x||_2^2}$$

where the notation $||(\ )||_p$ refers to the $p$th norm. We notice that the minimization is taken over all vectors $s$ with $K$ nonzero entries. The $K$-compressibility level equals 1 if the vector is exactly $K$-compressible, and falls in the range $[\frac{K}{N}, 1]$. The expected value of the $K$-compressibility level across the ensemble $X$, which we denote by $E[L]$, is also of interest as a holistic measure of compressibility. Similarly, the minimum value of the K-compressibility level across the ensemble serves as a metric for compressibility. In our development, we often plot the $K$-compressibility level as a function of the sparsity $K$ to study whether the data can be represented tersely in the basis of interest.

The $K$-compressibility level $L$ is straightforward to compute, provided that the basis $F$ is orthogonal (as assumed). To compute $L,$ first the vector $\hat{s} = F^{-1}x$ is determined; since the inverse of an orthogonal matrix is its transpose, this vector can be more easily calculated as $\hat{s} = F^T x$. The vector $\hat{s}$ indicates the component in each basis direction when the scene vector is expressed in the full basis of interest. Then, a vector $s^*$ is formed by maintaining the $K$ entries with largest magnitude in $\hat{s}$, and setting the remaining entries to 0. The $K$-compressibility level is given by $L = \frac{||s^*||_2^2}{||\hat{s}||_2^2}$. The approximation of $x$ which achieves the compressibility level, i.e. the $K$-sparse approximation which captures the largest fraction of the energy in the vector, is given by $x^* = Fs^*$. We call this the optimal $K$-sparse approximation.

III. COMPRESSING WEATHER DATA: APPROACH

Our method for developing compressive representations for aviation-relevant weather data is described. The method comprises the following steps: a) pre-processing of METAR data to obtain a weather scene; b) construction of the graph-spectral basis; and c) compressive representation and compressibility analysis in the graph-spectral basis. Each step is described.

*III.A. Weather-Scene Preprocessing*

In this study, compression of Meteorological Aerodrome Reports or METARs, which are weather reports from airports and permanent weather stations that are used for aviation forecasting/planning, was pursued. METAR data was obtained from the Aviation Weather Center [23]. Reports from locations within the contiguous United States (between $25.1° - 45.4°$ N, and $124.8° - 66.9°$ W) collected during a one-hour period were used to define a scene. Specifically, the recorded values of a weather quantity of interest from across this region were placed in a vector. In this initial study, we primarily focused on three weather quantities: temperature, flight categories (e.g. visual flight rule (VFR) vs. instrument flight rule (IFR)), and visibility. Along with the weather quantity of interest, the corresponding geographic locations (latitude, longitude) of the METAR sites were also tabulated. The locations of the METAR sites are shown in Figure 1. The focus in this initial study was on compression of several sample scenes of this sort from two time periods on January 18, 2021.

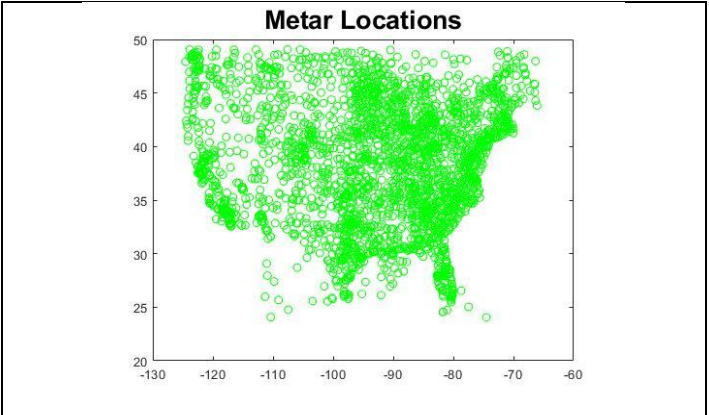

**Figure 1:** Compression is pursued for weather data scenes from METAR sites across the contiguous United States, shown here.

The weather vector was preprocessed in several ways. First, non-numerical quantities such as flight categories were mapped to numerical values. Also, spurious or absent data values were removed, and inconsistent reports were corrected (for example, varying maximum visibility distances were equalized). The vector after preprocessing was the scene vector which we sought to compress, and hence we denote it as $x$ from here on in keeping with the notation from Section II. For the data considered here, the weather scene vectors had between $N=2200$ and $N=2500$ entries.

*III.B. Constructing the Graph-Spectral Basis*

Aviation-related weather quantities, although constantly varying in time, exhibit spatial correlations and patterns. Thus, it is natural to pursue compression in bases that encode spatial distance. In our setting, developing such bases is complicated by the fact that the scene is irregularly sampled in spatial dimensions. For such data, a promising way to construct bases that reflect spatial interactions is to first define a graph which indicates spatially-close data locations, and then define the basis thereof [24-26].

In this work, a graph $G$ was formed from the geographical locations of the METAR sites, as follows. First, the graph was defined to have $N$ vertices labeled $1,...,N$, corresponding to the $N$ entries in the scene vector $x$, or equivalently the $N$ METAR sites represented in the scene. Then, an edge was drawn between two vertices $i$ and $j$, if and only if the geographical distance between the two vertex sites was less than a threshold $T$. In our analyses, we used a threshold of $T=70$mi, or approximately 1 degree in latitude.

We considered compression of the weather scene vector in the *graph-spectral* or *Laplacian-eigenvector* basis [24-26], which is defined from the *Laplacian matrix* of the graph $G$. The $NxN$ Laplacian matrix $H$ is constructed from the graph $G$ as follows: the entry at row $i$ and column $j$ is set to -1 if there is an edge between vertices $i$ and $j$, and is set to 0 otherwise. Then, each diagonal entry is selected so the corresponding row (and column) of $H$ sums to 0, i.e. the entry is selected as the negative of the sum of the off-diagonal entries.

The graph-spectral basis is specified by the eigenvectors of the Laplacian matrix $H$ (hence it is also known as the Laplacian-eigenvector basis). Formally, from properties of Laplacian matrices, $H$ can be written in the form $H = V\Lambda V^{-1}$, where $\Lambda$ is a diagonal matrix whose diagonal entries are the eigenvalues of $H$ and satisfy $0 = \lambda_1 \leq \lambda_2 \leq \cdots \leq \lambda_N$; and $V = [v_1, ..., v_N]$ is an orthogonal matrix containing the corresponding eigenvector. The matrix of Laplacian eigenvectors $V$, known as the graph-spectral basis, is the basis in which we considered compression. Thus, we defined the basis matrix $F$ for compression as $F=V$. The basis can be re-used for compression of multiple scenes (e.g., for different weather quantities and time periods), provided that the same METAR locations are used to form the scene vectors for these cases.

The basis vectors in the graph-spectral basis have interpretations related to the topology of the graph, which in our case captures geographical closeness of METAR sites. In particular, the eigenvectors contained in the basis are known to represent spatial waves of different frequencies, with the frequency/extent of the waves delineated by the corresponding eigenvalues. Thus, the eigenvectors associated with small eigenvalues are "slow" spatial waves which arise in data that have long-range correlations, while the eigenvectors associated with larger eigenvalues can capture higher-frequency and more localized characteristics. We note that the graph-spectral basis reduces to the two-dimensional Fourier basis in the case that the sampling locations form a grid with equal spacings. Thus, the graph-spectral basis is an analog for a Fourier basis for irregularly sampled scenes.

*III.C. Compressive Representations and Compressibility*

The compressibility of the weather scene vectors in the graph spectral basis is examined, and reconstruction of the scenes from compressed representations is also undertaken. Compressibility is analyzed using the definitions and techniques described in Section II. For each scene vector, the analysis of compression is primarily focused on three aspects:

1) The $K$-compressibility level is plotted as a function of the sparsity $K$, to understand how many basis vectors are needed for effective compression.
2) The dominant basis vectors, i.e. the basis vectors which have largest components in the scene, are visualized on top of the map of METAR sites.
3) The scene vector is reconstructed from a sparse representation, and the reconstruction is compared with the original.

## IV. COMPRESSION ANALYSIS: RESULTS

Compression was considered for METAR-based scenes of three weather quantities: surface temperature, flight rule conditions (e.g., visual flight rules vs instrument flight rules), and visibility. Results are shown for scenes from two different time periods on January 18, 2021.

*IV.A. Surface Temperature Data*

Surface and higher-altitude temperatures are of interest to aviation-system stakeholders, because they impact aircraft performance (e.g. takeoff performance at high temperatures) and indicate the possibility of certain hazards such as ice accretion. Here, compression of surface-temperature scene vectors derived from METAR data was considered. The scenes contained data from 2377 METAR sites. Figure 2 shows the $K$-compressibility levels vs the sparsity $K$ for a particular temperature scene vector. It is seen that 90% of the scene-vector energy is captured using only 10 basis vectors, with 93% captured using 50 basis vectors.

Next, in Figure 3, six dominant basis vectors are visualized on top of the map of METAR sites, for this temperature scene vector. For each basis vector, entries with sufficiently-large magnitude are colored as either blue or black on the corresponding METAR site on the map. The basis vectors distill key patterns in the temperature profile: for instance, the first basis vector indicates an offset in the temperature from 0-degrees C, the second basis vector shows a gradient between the Eastern and Western United States, and the third basis vector captures anomalies in the Northern plains, Northern New England, and Florida.

In Figure 4, the absolute error in a reconstruction of the scene from a 50-sparse approximation is shown. The temperature is accurate to within 2 degrees at a 51% of the METAR sites, and to within 5 degrees at 90% of the METAR sites. As might be expected, the errors are somewhat larger at locations with anomalous temperatures, such as at some locations in the Northern plains.

Finally, it is instructive to visualize the dominant basis vectors for a second surface-temperature scene vector, obtained from a different time period. The six dominant basis vectors for the second scene are plotted on top of the METAR sites map in Figure 5. The plots show that some dominant basis vectors are shared between the two scenes, however some of the dominant basis vectors also differ: for example, the West-to-East gradient basis vector is not a dominant vector for the second scene, but an alternate vector that identifies variations across the Mountain West is present. The variation in the dominant basis vectors highlights the value of considering scene-dependent terse descriptions of weather data, rather than a fixed low-dimensional basis. Each scene may be sparse in a basis of interest (in this case the graph spectral basis), however the dominant vectors in the sparse approximation may change. In this case, the variation in the dominant vectors is relatively slight, because the two time periods are only a few hours apart and hence have similar temperature profiles. We anticipate that the profiles may vary much more significantly between surface temperature scenes with days or weeks of separation.

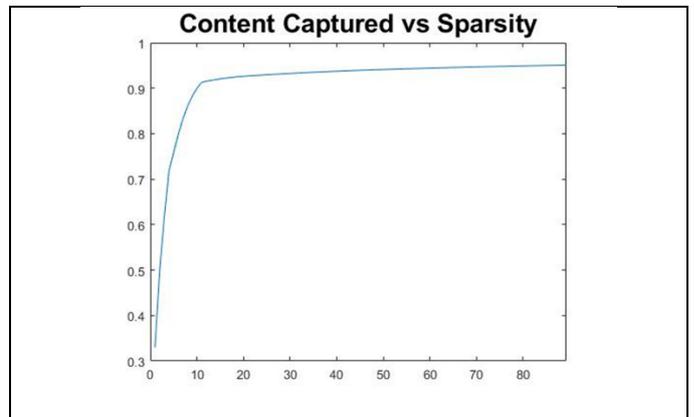

**Figure 2:** The $K$-compressibility level is plotted as a function of $K$, for a surface temperature scene vector with $N=2377$ elements. 90% of the energy is captured using only 10 basis vectors.

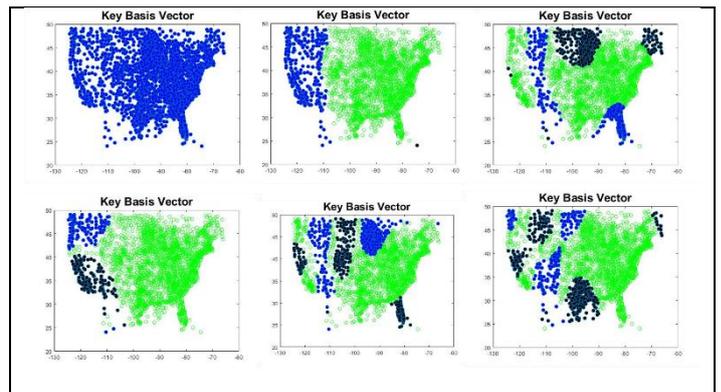

**Figure 3:** Six dominant basis vectors for the temperature scene vector are visualized on the METAR sites map. The basis vectors are ordered according to the magnitude of their component in the scene. They are displayed from left to right across the top row, and then across the bottom row.

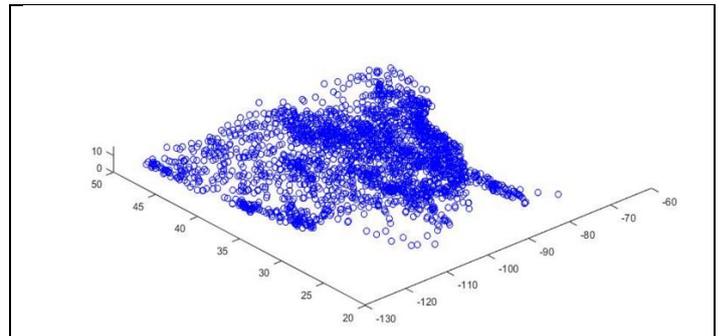

**Figure 4:** The absolute error in a reconstruction using a 50-sparse approximation is shown on top of the map of METAR sites. The reconstruction is accurate to within 2 degrees at over 51% of the sites, and to within 5 degrees at over 90% of the sites.

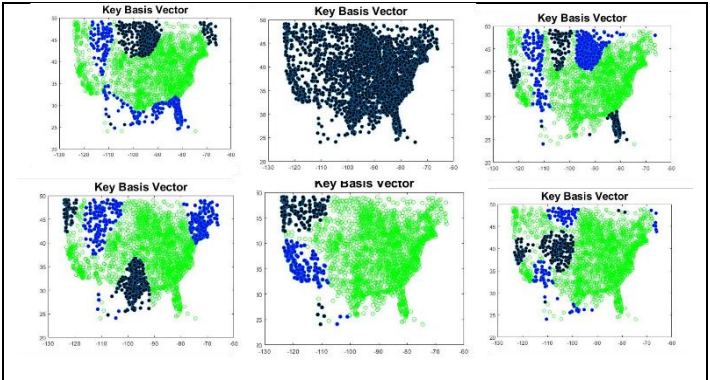

**Figure 5:** Six dominant basis vectors are visualized on the METAR sites map, for a second temperature scene vector. The basis vectors are ordered according to the magnitude of their component in the scene.

x

*IV.B. Flight Categories Data*

Compressibility of flight categories data at METAR sites in the graph spectral basis was also examined. Since flight categories at terminals determine arrival and departure rates as well as spacing in the air, they directly impact airspace resource capacities and hence are of concern for traffic managers in the design of TFM. To develop compressive representations of flight categories data, we first translated the data to a numeric form. In this initial study, we indicated the visual flight rules (VFR) category as 0, and other flight rule categories (including MVFR, IFR, and LIFR) as 1.

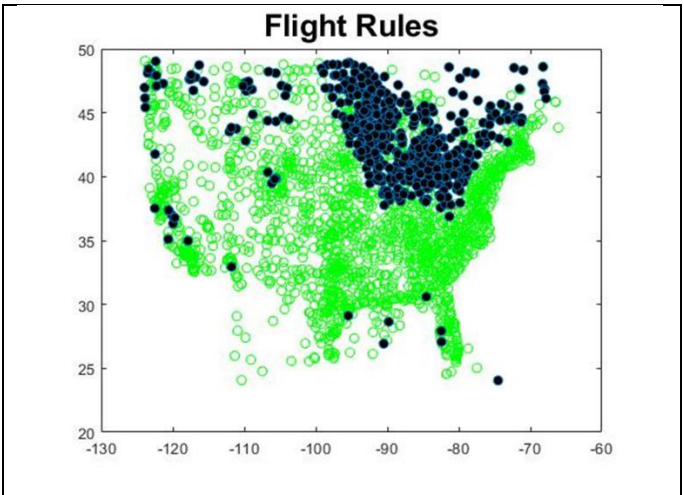

**Figure 6:** A flight categories scene vector obtained from METAR data is shown. Green circles indicate METAR sites with VFR conditions, while black circles indicate sites with non-VFR conditions.

A flight-categories scene vector for a time point on January 18$^{th}$ was considered. The vector had $N$=2271 entries. The flight-categories scene vector is shown in Figure 6: non-VFR conditions are prevalent in the upper Midwest and the interior Northeast, with more isolated pockets of non-VFR conditions in other parts of the contiguous United States. In Figure 7, the $K$-compressibility levels are shown as a function of the sparsity $K$. In this case, about 60% of the scene-vector's energy is captured using 50 basis vectors, and 75% is captured using 100 basis vectors. The plot shows that the flight categories scene is also highly compressible, although to a somewhat lesser degree than the surface temperature scene. In Figure 8, the four dominant basis vectors are shown: these basis vectors are amplified at different points in the Northern tier of the contiguous United States, and in parts of the Western United States, reflecting strong gradients in the flight categories in these areas. In Figure 9, reconstruction of the flight categories scene vector from a 50-sparse approximation is illustrated. Noting that a direct reconstruction yields fractional flight-rule values, we scaled the direct reconstruction to match the energy of the original scene vector and then rounded the estimate to obtain the reconstruction; the result is shown in the figure. The reconstruction is found to be quite accurate, correctly correctly identifying about 88% of the non-VFR locations, and about 93% of all locations' flight categories.

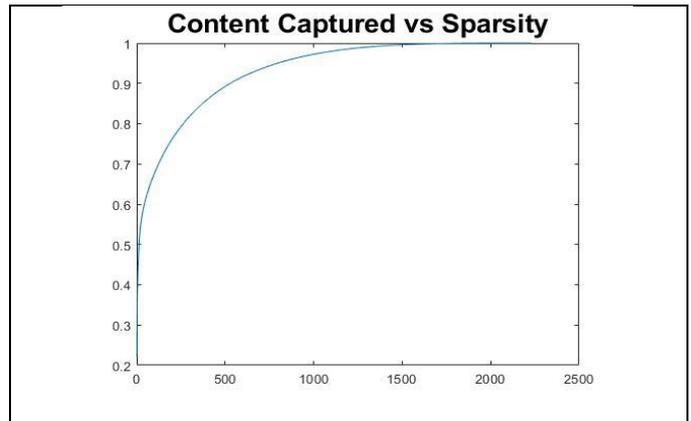

**Figure 7:** $K$-compressibility levels are shown as a function of $K$, for the flight-categories scene vector.

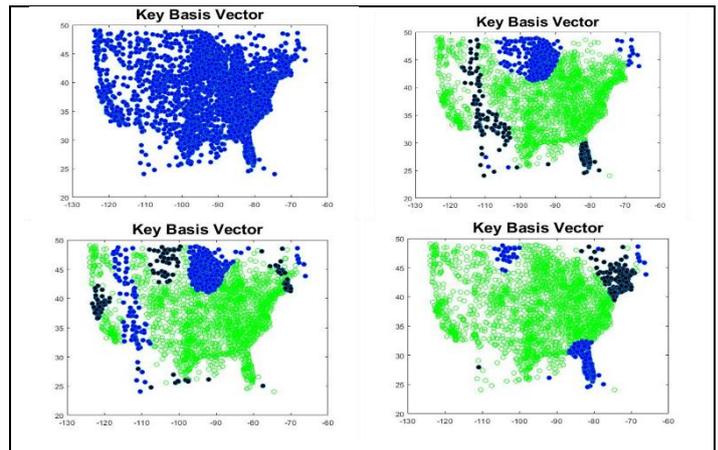

**Figure 8:** Four dominant basis vectors are visualized on the METAR sites map, for the flight categories scene vector. The basis vectors are ordered according to the magnitude of their component in the scene (from top left to bottom right in the figure).

Compression of the flight categories data was also undertaken for a second time point. The flight-categories scene, dominant basis vectors, and reconstructions using 50 and 150 basis vectors are shown in Figure 10. In this case, the flight categories scene at the second time point differs significantly from the first scene, with significant areas with instrument flight rules in Texas, and comparatively sparser prevalence of non-VFR conditions in the upper Midwest. The dominant basis vectors reflect the different structure of the scene, with several basis vectors having amplified entries in Texas and the Southwest. Again, effective reconstruction is achieved using 50 basis vectors, with a larger number of basis vectors (150) serving to sharpen the resolution of the non-VFR areas.

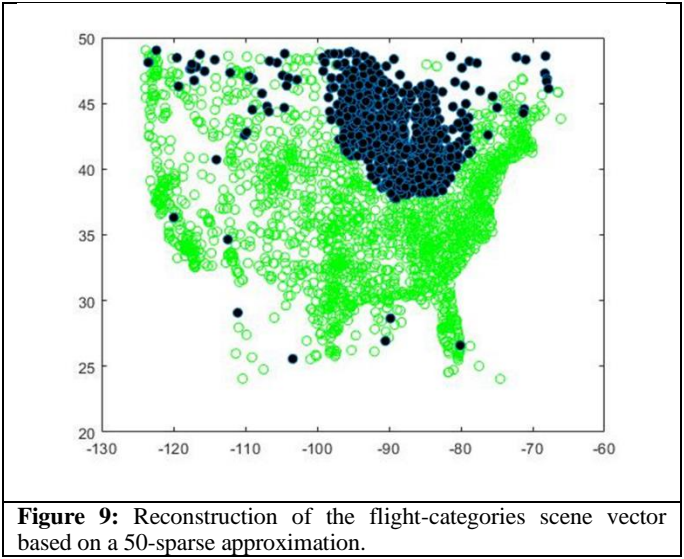

**Figure 9:** Reconstruction of the flight-categories scene vector based on a 50-sparse approximation.

*IV.C. Visibility Reduction Data*

Visibility measurements and forecasts are also used for TFM. Visibilities in the terminal area and en route influence airspace resource capacities, both through their role in deciding flight rules, and through their direct impact on traffic control (e.g., due to altered spacing requirements, additional workload for controllers, etc). Thus, strategic TFM requires consideration of visibilities across the airspace system. With this motivation, we pursued compression of visibility data extracted from METARs. To develop a scene vector, recorded visibilities from METAR sites across the contiguous United States during a 1-hour time period were first tabulated. The data was pre-processed to equalize the maximum visibility at 10 mi. Then, the scene vector was defined as the reduction in the visibility from the maximum. The visibility reduction rather than the visibility itself was used, to clarify the compression technique's ability to capture areas of reduced visibility which are of concern in TFM.

The scene vector was constructed and compression in the graph spectral basis was undertaken, for the same time period as was considered for the first flight-categories scene. The $K$-compressibility level is shown as a function of $K$ for this scene in Figure 11. Next, in Figure 12, the four dominant basis vectors represented in the scene are shown. These results for the visibility-reduction scene show a close parallel with the flight-categories scene vector, as might be expected given the interdependence of the two quantities. In particular, the $K$-compressibility level plot is similar in shape to that for the flight-categories data, and the dominant basis vectors are also similar. The results indicate that visibility data is also compressible in the graph-spectral basis, with the dominant basis vectors giving insight into the visibility-reduction profile.

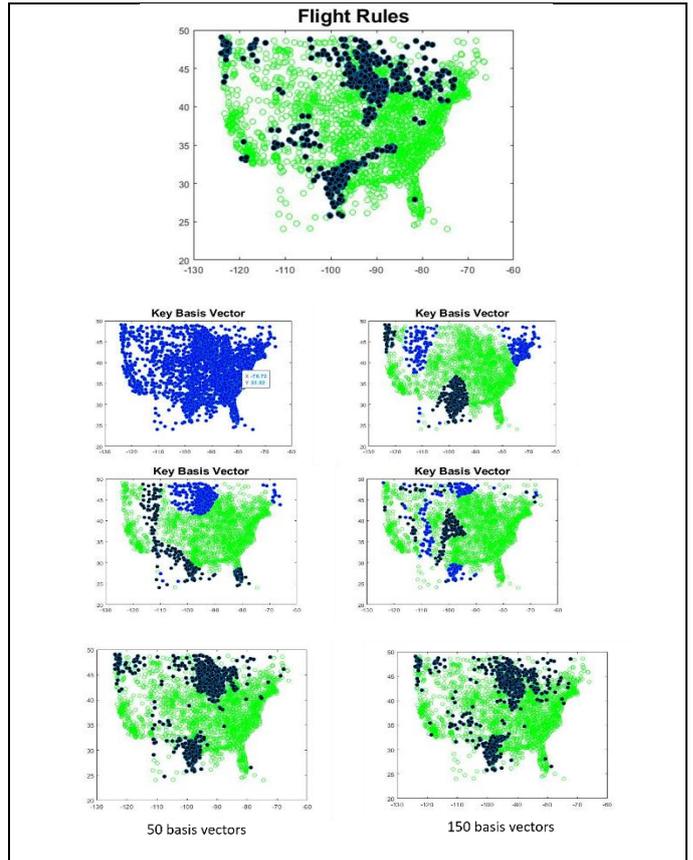

**Figure 10:** Compressive representation of a second flight-categories scene is considered. The scene, four dominant basis vectors, and reconstructions using 50 and 150 basis vectors, are shown from top to bottom.

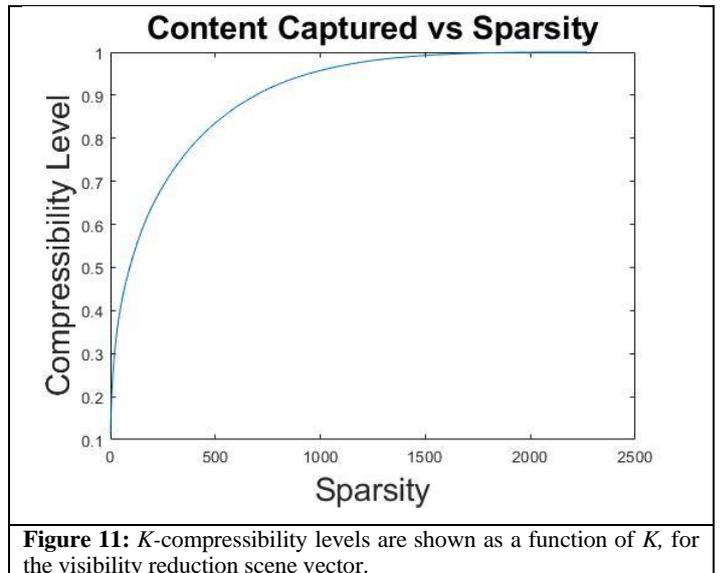

**Figure 11:** $K$-compressibility levels are shown as a function of $K$, for the visibility reduction scene vector.

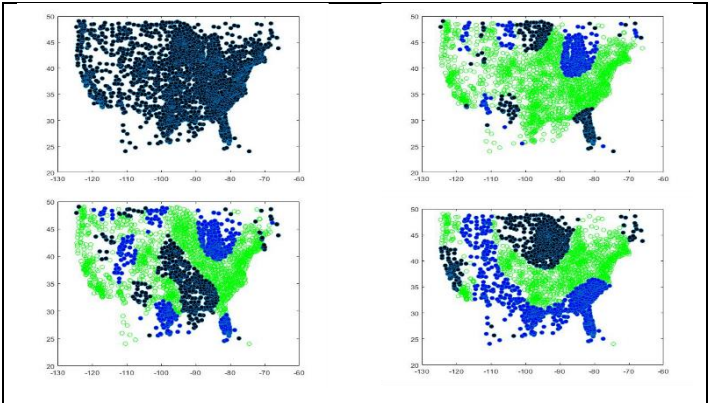

**Figure 12:** Four dominant basis vectors are visualized on the METAR sites map, for the visibility reduction scene vector. The basis vectors are ordered according to the magnitude of their component in the scene (from top left to bottom right in the figure).

V. DISCUSSION AND IMPLICATIONS FOR TFM DESIGN

The main focus of our study was to examine whether aviation-relevant weather data at the national scale, specifically METAR data, is compressible in the graph-spectral basis. The data analyses in the previous section give a preliminary indication that weather quantities of interest are in fact compressible, in that each scene is representable by a small number of basis vectors which however vary among the scenes. Our case studies indicate that a significant fraction of the scene vector's content can be captured using a small fraction of the graph-spectral basis vectors (e.g., 50 of about 2500 or 2% of the basis vectors). Also, the case studies show that the dominant basis vectors capture key spatial properties of the weather data, and also that a sparse approximation is sufficient for accurate reconstruction of the scene vector. Holistically, our studies suggest that the compression framework can provide terse representations for aviation-relevant weather and weather impact data, which are both computationally attractive and able to distill salient spatial features of the weather. In this section, as a starting point toward further work, we discuss at a conceptual level how compressive representations may be used for designing strategic TFM. Specifically, we postulate three possible uses of the compressive representations:

*TMI Selection Based on Dominant Basis Vectors.* Initial selection of traffic management initiatives across the United States National Airspace System on a day of interest is challenging, because of the sophisticated spatiotemporal pattern of weather impacts and the uncertainty inherent to weather at strategic look-ahead times. Previous data-theoretic efforts on TMI selection have focused on classifying historical weather days, and comparing the forecasted weather on the day of interest to the historical categories [9,16,17]. This study gives a slightly different perspective, that a weather scene can be represented by a varying but small set of basis vectors, which distill key spatial features of the weather. This structure may enable selection of TMIs based on the (forecasted) dominant basis vectors for the weather scene on the day of interest. Specifically, we anticipate that the identified dominant basis vectors can be mapped to a set of preferred TMIs, either through tabulation of historical days with these dominant basis vectors, or through model-driven design given sparsely-structured weather. In particular, noticing that the weather has a sparse basis, it may be possible to undertake TMI selection and design for a relatively small number of canonical weather scenes, and then select TMIs on a day of interest based on these cases.

*Effective Learning of TFM Strategies.* The aviation community has been increasingly interested in leveraging data-driven learning to support traffic management, including in designing TFM strategies [27,28]. Regardless of the specific approach taken, learning of models or decision strategies requires terse representations of system inputs and parameters. In the TFM setting, terse representation of weather scenes is necessary for robust learning without overfitting. Compression in the graph-spectral basis allows for a sparse representation, in that it distills a weather scene vector into a small number of key basis vectors and associated indices or eigenvalues. Thus, we expect that compression of weather data may support efficient learning of air traffic models and traffic flow management strategies.

*Identification of Bellwether Data/Forecast Variables.* Because weather data and forecasts are high dimensional, operators often benefit from focusing on a subset of bellwether data and forecast variables. The compressive sensing literature clarifies that compressibility of a high dimensional data scene allows reconstruction from a much sparser sampling of the scene [18-22]. Thus, compressibility allows for definition of a subset of bellwether data or forecast variables (e.g., data from a subset of METAR sites), which fully characterize the weather scene and hence are sufficient for TFM design. We anticipate using compressibility to select such bellwether variables, both to simplify computation of TFM strategies and to give operators easier insight into weather characteristics.

We expect to develop these uses for compressive representations in TFM design in future work.

ACKNOWLEDGMENT

The author thanks Drs. C. Taylor and E. Vargo of the MITRE Corporation for their insights regarding this work. The author is grateful for the support provided by the U.S. National Science Foundation through an Intergovernmental Personnel Act agreement with Washington State University.